# Can ChatGPT be Your Personal Medical Assistant?

Md. Rafiul Biswas
*College of Science and Engineering*
*Hamad Bin Khalifa University*
Doha, Qatar
0000-0002-5145-1990

Ashhadul Islam
*College of Science and Engineering*
*Hamad Bin Khalifa University*
Doha, Qatar
0000-0002-9717-3252

Zubair Shah
*College of Science and Engineering*
*Hamad Bin Khalifa University*
Doha, Qatar
0000-0001-7389-3274

Wajdi Zaghouani
*College of Humanities and Social Sciences*
*Hamad Bin Khalifa University*
Doha, Qatar
0000-0003-1521-5568

Samir Brahim Belhaouari
*College of Science and Engineering*
*Hamad Bin Khalifa University*
Doha, Qatar
0000-0003-2336-0490

*Abstract*—The advanced large language model (LLM) ChatGPT has shown its potential in different domains and remains unbeaten due to its characteristics compared to other LLMs. This study aims to evaluate the potential of using a fine-tuned ChatGPT model as a personal medical assistant in the Arabic language. To do so, this study uses publicly available online questions and answering datasets in Arabic language. There are almost 430K questions and answers for 20 disease-specific categories. GPT-3.5-turbo model was fine-tuned with a portion of this dataset. The performance of this fine-tuned model was evaluated through automated and human evaluation. The automated evaluations include perplexity, coherence, similarity, and token count. Native Arabic speakers with medical knowledge evaluated the generated text by calculating relevance, accuracy, precision, logic, and originality. The overall result shows that ChatGPT has a bright future in medical assistance.

*Keywords—ChatGPT, fine-tuning, medical bot, LLM, Arabic*

## I. INTRODUCTION

A chatbot is designed to communicate with users, either through text or spoken language, often with the goal of providing information, answering questions, or assisting with tasks [1]. In the context of the Arabic language, chatbots have proven to be valuable tools by allowing users to interact in their preferred language. Arabic chatbots can assist in various sectors, including customer service, education, and marketing, by providing information, answering queries, and offering support [2]. However, it requires the processing of large amounts of Arabic corpora for effective output, and the dialect of the Arabic language varies from country to country, which is a challenge to build an Arabic chatbot. Also, there is a limited representation of Arabic chatbots targeting healthcare professionals primarily involved in tasks such as patient intake automation and aiding in triage and diagnosis [3]. Moreover, healthcare data is highly sensitive, and ensuring robust security measures to protect patient privacy is paramount. Therefore, the creation of healthcare-focused chatbots demands a careful and comprehensive approach to address privacy issues effectively.

ChatGPT, the latest innovation by OpenAI in Large Language Model (LLM), has emerged as a transformative tool across diverse domains [4]. The continuous development of the LLM model has extended the capabilities of ChatGPT far beyond its initial design, allowing users to enter a new era of AI-powered collaboration and innovation across a spectrum of industries and disciplines. ChatGPT can be used for different purposes, such as text generation, image generation, and audio file conversation [5]. Through its adept understanding of human language and context across multiple languages, ChatGPT leverages a wide array of use cases across various sectors, redefining communication, problem-solving, and engagement. Even though there are other LLMs, such as Google Bard, Meta, Llama2, and so on, ChatGPT has been unbeaten due to its human-like responses in more than 50 languages [6]. In contrast to earlier chatbots, ChatGPT possesses the ability to recall the user's previous statements during a conversation, facilitating seamless dialogue [7]. Moreover, ChatGPT has undergone significant enhancements, allowing users to input both text and visual images concurrently. The recent release of GPT turbo-3.5 has enabled the developer and researcher to fine-tune the GPT 3.5 model [8]. The fine-tuned model will optimize the ChatGPT for specific tasks by enhancing its efficiency and ensuring accuracy. For example, fine-tuned models have the potential to bring significant improvements to various domains such as customer service, medical diagnoses, content writing, aid in education, and ethical decision-making [9], [10]. It's important to note that while Turbo GPT-3.5 can be a valuable tool in medical chatbots, it should complement, not replace, the expertise of trained healthcare professionals.

This study explores the capability of fine-tuned LLM on medical datasets in the Arabic language. It uses the extensive Arabic Healthcare Question & Answering (Q&A) dataset (MAQA), comprising over 430k questions across 20 medical specializations [11]. This is the largest collection for medical Q&A in the Arabic language so far and is used as a medical bot [12]. We have used the MAQA dataset for test purposes, which allows us to explore the capability of ChatGPT by engaging with different datasets and different languages. The rest of the paper is organized as follows. Section 2 describes the dataset. Section 3 describes the methodology. Section 4 describes the results. Section 5 elaborates on the discussions through outcome analysis and limitations.

## II. DATASET DESCRIPTION

The dataset contains over 430,000 questions distributed across 20 medical specializations. The questions were collected from various online websites. The data is stored in an SQLite database and is publicly available [11]. The MAQA corpus is a substantial collection of healthcare-related Q&A in the Arabic language, suitable for tasks like bot development, question answering, text classification, and word embedding models. There are six columns for each data: q_body, a_body, q_body_count, a_body_count, category, category_id. The q_body is the questions asked by patients, and the a_body is the answer provided by the healthcare professionals. The questions and answers both are in Arabic. Table 1 describes dataset attributes, and Figure 1 shows the number of Q&A per disease contained in the dataset.

TABLE I. DATASET DESCRIPTION

| Column | Paraphrased in English |
|---|---|
| q_body: questions asked by patients | My sister was born and lost the fetus. What do you do with the milk? Does she express or leave it? And if she leaves it, will it affect her in the long run with breast cancer? |
| a_body: answer provided by doctor | Treatment must be taken to stop the milk, and pumping must be done until it stops |
| q_body_count: word counts | 21 |
| a_body_count: word counts | 9 |
| category: disease category | Gynecological |
| category_id: disease id | 15 |

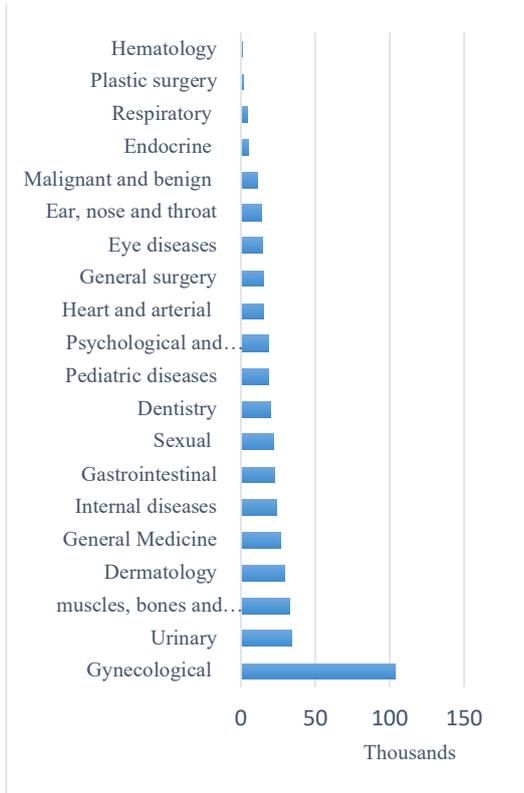

Fig. 1. Number of Q&A for each disease

### A. Ethical Consideration

The use of conversational AI for healthcare raises important ethical considerations. Preserving patient privacy is of utmost importance, requiring the anonymization of data and the implementation of secure access controls. It also requires consent patient from before collecting data or using it for conversations. In the context of this research, the dataset was sourced from publicly available websites [11], with no patient information disclosed, and so no consent was obtained for its usage. Therefore, ethical considerations were not applicable, as the results do not expose patients' identities.

## III. METHODOLOGY

### A. Preparation of Data for Finetuning

The original dataset remained in SQL database format. We extracted the data from the database to a CSV file. We cleaned the dataset by keeping only Arabic text and removing incomplete questions and answering. To fine-tune the gpt-3.5-turbo model, it requires preparing data in JSONL format. Each JSON message has three roles: system, users, and assistant. An example of a JSONL message is shown in Figure 2.

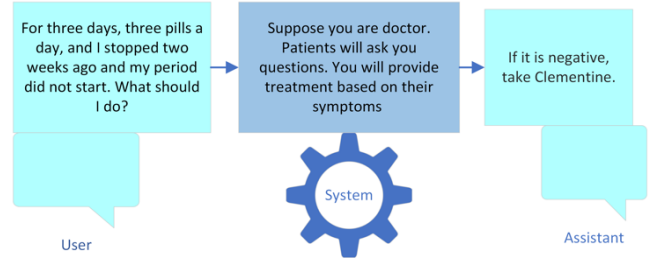

Fig 2. JSONL file

### B. Prompt Engineering

Prompt engineering bears significant importance in getting optimized results, particularly in the medical sector, as it involves deliberately designing input prompts to guide the model toward desired outputs [13]. This process allows for the creation of realistic patient scenarios, the generation of diverse multiple-choice questions, and explanations of complex medical concepts. We designed three prompts: 'system', 'user' and 'assistant'. The 'system' prompt is used to guide how the chatbot will function. The 'user' prompt is used to ask questions to 'assistant' prompt. The 'assistant' prompt provides answers to 'user' prompt. We used 'q_body' as the user prompt and 'a_body' as the assistant prompt. Figure 3 shows an example of the prompt.

Fig. 3. Example of Prompt

We evaluated the prompt by comparing the message body of the assistant prompt with the original answer. The answers generated by the fine-tuned model were specific to the disease and similar to the original answer. However, we found that answers generated without a fine-tuned model were generic replies for disease and descriptive analysis. This experiment was performed as a test case on the small data set, expecting that it would reflect on the large and diverse dataset.

### C. File Upload

We chose Gynecological disease (category 15) for the experiment because it is the largest sample, having 103,683 Q & A. We randomly selected 4000 Q&A for training purposes and 1000 Q&A for validation purposes from Gynecological disease. We uploaded JSONL files to OpenAI's file endpoint along with the OpenAI API key in the system environment variables for authentication. The larger the number of training datasets, the better performance of the fine-tuned model. However, fine-tuning models with large datasets increases the token size. Eventually, it will be expensive for larger sets of tokens.

### D. Model Fine-tuning

Fine-tuning GPT models can improve model performance for specific tasks, but designing effective prompts requires time and effort, breaking tasks into smaller steps, and measuring accuracy [15]. Initially, we uploaded preprocessed training data using the OpenAI API key to the GPT-turbo-3.5. Then, we created a fine-tuned job ID by making the API call.

We provided the fine-tuned IDs for the train and validation data files. Once the job is finished, we provide a suffix for the output model name. Later, we used this fine-tuned model for question and answering. Once fine-tuning is completed, we find the job results. The training job will automatically upload a result file named step_metrics.csv, which contains training and test loss and accuracy for each training step. However, training with the GPT model is costly and depends on the number of tokens. The complete code used in this study can be accessed at the following link: https://github.com/rafiulbiswas/Fine-tuned-ChatGPT-model.git.

## IV. RESULTS AND EVALUATION

The OpenAI API allows for training using both training and validation datasets, offering loss figures for both throughout the training process [14]. These metrics serve as an initial indicator of the model's progress, especially when contrasted with outputs from the base model on equivalent test conversations. The model was trained for three epochs, and the recorded training loss of 0.3355 is shown in Figure 4. It implies that the training loss was going down with more data and the number of epochs. To evaluate the performance of fine-tuned GPT turbo 3.5, we performed two analyses: automated and human evaluation.

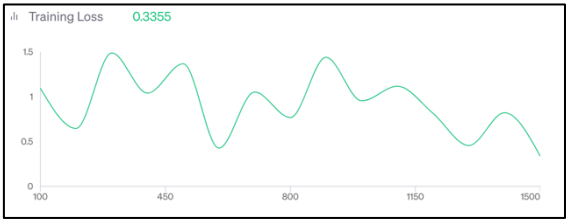

Fig. 4. Training Loss

### A. Automated Evaluation

We compared the generated answer with the original answer. We observed the Perplexity score, ROUGHE score, Coherence score, and Count of the token.

**Perplexity Score**: Perplexity score is widely used in natural language processing (NLP) and machine learning to evaluate the performance of language models on the generalization of text based on the training dataset. It quantifies how well a probability distribution or probability model to predict unfamiliar or previously unseen data [16]. Mathematically Perplexity Score is calculated as:

$$PP(W) = 2^{-\frac{1}{N}\sum_{i=1}^{N} \log_2 P(w_i, w_2, w_3, \ldots, w_{i-1})} \qquad (1)$$

- $PP(W) = perplexity\ on\ a\ dataset\ W$.
- N = total number of words in the dataset
- $w_{i=}$ i-th word in the dataset
- P(wi|w1,w2,...,wi−1) = probability assigned model

A lower perplexity score indicates that the model's predictions are confident and precise, whereas a higher perplexity score suggests that the model's predictions are uncertain and less accurate. Figure 5 shows the perplexity score obtained for the test dataset. The average perplexity score was measured 13.96 which signifies model is moderately confident and precise to the uncertainty in predicting each word in a sequence.

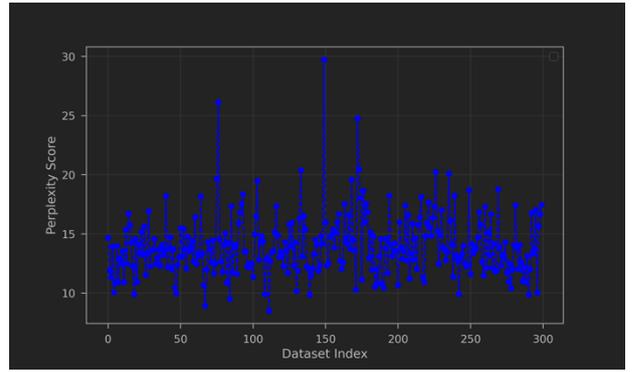

Fig.5. Perplexity Score

**Similarity Score**: The similarity score compares content, meaning, or characteristics between two pieces of text [17]. A higher similarity score implies increased proximity to the original text. In the medical domain, a higher similarity score is preferred. There are different ways to measure similarity scores such as cosine similarity, Jaccard similarity score, TF-IDF similarity, and others. We performed cosine similarity between the original answer and the generated answer. In the context of NLP, cosine similarity is often used to measure the similarity between vectors representing the distribution of words in a given space (such as word embeddings). The formula for computing the cosine similarity between two vectors, A and B, is as follows:

$$\text{cosine similarity}(A, B) = \frac{A.B}{||A||.||B||} \qquad (2)$$

Figure 6 depicts cosine similarity. The fine-tuned model average similarity score was measured at 0.10. A cosine similarity score of 0.10 suggests a relatively low level of similarity on average between the vectors. This could imply that the representations generated by the fine-tuned model are not very similar across the dataset or context being considered.

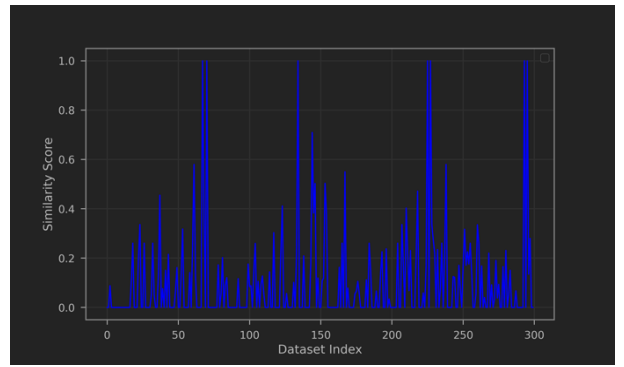

Fig. 6. Similarity score

**Coherence Score:** Coherence score is used to measure how well the text in a topic relates to each other [18]. Coherence scores 0 (low coherence) and 1 (high coherence) are relative measures used to assess the quality of topics within a topic model. A higher coherence score represents the model's ability to generate consistent, clear, and relevant text. The average coherence score for the test dataset was measured at 0.3336. This score indicates the coherence or interpretability of the topics generated by the model, and it is not a direct measure of the percentage of texts related to the topic. A coherence score of 0.3336 suggests a moderate level of topic coherence. This means that there is a reasonable degree of semantic similarity or connectedness among the terms used to describe the answer.

**Fine-tuned Model vs. Core GPT 3.5 Model**: We generated text for the same question by using the GPT 3.5 model and fine-tuned model. Notably, it's evident that the GPT 3.5 model tends to produce a higher count (50 to 200) of tokens compared to the original answer (0 to 50). Interestingly, the number of tokens generated by the fine-tuned model is closer to the original answer. This states that the fine-tuned model is learning and can perform better if the training datasets are sufficient. Figure 7 shows the comparative analysis.

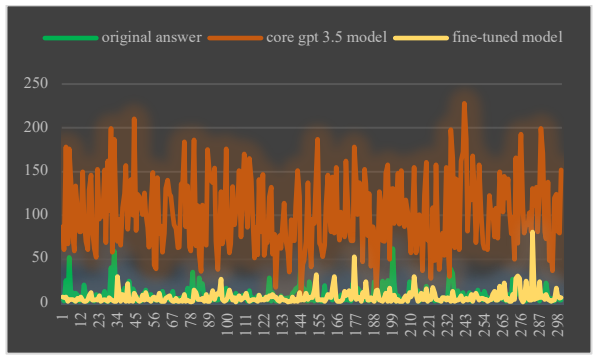

Fig.7. GPT 3.5 vs fine-tuned model

*B. Human Evaluation*

To evaluate the performance of the fine-tuned model, we took help from two medical professionals who were native Arabic speakers. They labeled 500 generated text based on the following criteria. Each generated text can be labeled +1 (very bad), +2 (bad), +3 (acceptable), +4 (good) and +5 (very good). Each generated text was compared to its original answer. We measured the inter-annotator's agreement on the labeled data using Cohen's Kappa. We compared all the pairwise combinations of the annotator's agreement and found the average of the result, which was 0.6377. We followed the study [19] for generating benchmark criteria for the generated text. Table 2 shows the descriptive analysis.

TABLE II. DESCRIPTIVE ANALYSIS

|  | Mode | Median | Mean | Std. dev |
|---|---|---|---|---|
| Relevance | 2.000 | 3.000 | 3.000 | 1.107 |
| Precision | 2.000 | 4.000 | 3.220 | 1.217 |
| Logic | 4.000 | 4.000 | 3.980 | 0.892 |
| Originality | 4.000 | 4.000 | 3.940 | 0.913 |
| Accuracy | 4.000 | 4.000 | 4.100 | 0.863 |

**Relevance:** It observes if the generated answer is relevant to the question. The mode of 2 (bad) for relevance scores indicates that much of the generated text by the fine-tuned model is not relevant to the original answer. The mean and median of 3 (acceptable) indicates that the average relevance score is acceptable. The standard deviation of 1.107 indicates that the values are somewhat spread out around the mean and indicates moderate agreement among the assessors regarding relevance.

**Accuracy:** It denotes if the generated answer belongs to accurate information and it can be verified through evidence or other means. Accuracy stands out with a mode, median, and mean score of 4 (good), indicating a high level of agreement among assessors regarding accuracy. The standard deviation (0.863) is relatively low, highlighting low variability in accuracy scores.

**Precision:** It describes if the generated answers are specific, precise, and unambiguous. It has a mode of 2 and a median of 4, suggesting a wide range of assessments. The mean precision score is 3.22, indicating a slightly negatively skewed distribution. The standard deviation (1.217) is relatively high, suggesting greater variability in precision scores compared to relevance.

**Originality:** It checks whether the generated text creates new insights and avoids repeated information. The mode of 4 (good) also indicates that 4 (good) is the most frequent value for the originality score. The median and mean are also both 4, indicating that the originality scores are very concentrated around 4. The standard deviation of 0.913 is similar to the standard deviation for logic, indicating that the originality scores are also not very spread out around the mean.

**Logic:** It shows if the conveyed idea adheres to coherent and logical reasoning. The mean, mode, and median of 4 (good) indicate that the generated text is good and keeps sound logical reasoning. It means the generated answer bears fruitful meaning to the questions. The standard deviation of 0.892 is relatively small, indicating that the values are not very spread out around the mean.

V. DISCUSSIONS

This study fine-tuned the GPT turbo 3.5 model on a comprehensive Arabic Healthcare Question & Answering (Q&A) dataset (MAQA) and reveal the promising capabilities of this fine-tuned model as a personalized medical assistant in the Arabic language. Here, we discuss the implications of these findings, emphasizing the model's potential in providing tailored medical information and assistance.

One of the standout observations from our study is the fine-tuned model's ability to generate responses that are remarkably close in token count to the original answers. This suggests a high degree of learning and adaptation, essential traits for a reliable medical assistant. When the model's training dataset is sufficiently rich and diverse, it demonstrates the capacity to understand and generate content that aligns closely with the authoritative medical information [20].

Our automated evaluation of the fine-tuned model includes various metrics such as perplexity, similarity score, and coherence score. The low perplexity score (13.96) indicates that the model excels at generalizing text based on its training data, enhancing its ability to provide precise and confident answers. Furthermore, the cosine similarity score, though modest (0.10), can be viewed as a positive sign, as it aligns with the expected low similarity in medical responses, where diversity is often encouraged to address a wide range of patient scenarios [21].

The human evaluation component of our study enlisted the expertise of native Arabic-speaking medical professionals. Notably, the model achieved high scores for accuracy, with a mode, median, and mean all at 4.0, reflecting a strong consensus among assessors regarding the model's capacity to provide accurate medical information. This finding underscores the potential of the fine-tuned model as a reliable and trustworthy medical resource. Although relevance and precision received slightly lower scores (mode of 2 for relevance and for precision), these attributes can be further fine-tuned to enhance the model's performance. The moderate standard deviations suggest that there is room for improvement, and additional training data or fine-tuning may lead to better alignment with user expectations. Moreover, the model has the ability to provide answers that are both logically sound and avoid repetition, a crucial aspect of medical consultations.

## VI. LIMITATIONS

While this study presents a potential analysis of using a fine-tuned model as a medical assistant, it has certain limitations. The training and evaluation datasets comprise online question-answer forums, which may not fully capture the nuances of real-world doctor-patient dialogues. Performance on direct clinical conversations remains untested. Many of the original answers in the dataset lack sufficient explanation, which impacts the performance of the fine-tuned model. Therefore, it is recommended to utilize a larger dataset sourced from hospitals to address these limitations effectively. The relatively small sample size for training (4000 examples) limits model exposure to diverse medical vocabulary and conversational patterns. More training data would be ideal. However, fine-tuning with a large dataset results in an increased number of tokens, leading to higher costs when using the OpenAI API. Additionally, there is a limitation on the number of tokens that can be used per training cycle. Consequently, while there is potential for ChatGPT to serve as a personalized medical assistant, questions arise regarding the practicality of using it extensively in the medical domain, especially in terms of resource requirements.

## VII. CONCLUSION

This preliminary study shows the potential of using fine-tuned LLMs for medical assistance applications in Arabic. Its performance in terms of accuracy and logical reasoning positions it as a valuable tool for providing trustworthy medical guidance. While there is room for further refinement, these findings emphasize the model's potential in offering personalized medical support and information to individuals, enhancing access to healthcare knowledge. Future work should focus on constructing high-quality conversational datasets from hospital and clinic interactions to better cover the nuances of professional medical dialogues. Additionally, ensemble approaches combining retrieval models with generative LLMs could boost overall performance. Rigorous testing on practical clinical deployments encompassing diverse medical conversations and user populations would be vital before large-scale adoption in real-world healthcare scenarios.

## ACKNOWLEDGMENT

We acknowledge the voluntary work for data labeling by two native Arabic speakers Bouthaina Gasmi and Hagar Mohamed Hussein. This publication was partially funded by NPRP grants 14C-0916-210015 and NPRP13S-0206-200281 from the Qatar National Research Fund (a member of Qatar Foundation). The findings herein are solely the responsibility of the authors.